\documentclass{article}
\usepackage[margin=1in]{geometry}
\usepackage[utf8]{inputenc}
\usepackage{algorithm}
\usepackage[noend]{algpseudocode}
\usepackage{amsmath,amssymb,amsfonts,amsthm}
\usepackage[numbers]{natbib}
\usepackage{hyperref}
\usepackage{booktabs}

\usepackage{siunitx}

\usepackage{xcolor}

\title{Estimation for Quadrotors}
\author{\parbox{0.3\linewidth}{\centering Stefanie Tellex\\Brown University}
  \parbox{0.3\linewidth}{\centering Andy Brown\\Udacity, Inc.}
\parbox{0.3\linewidth}{\centering Sergei Lupashin\\Fotokite}    
}

\begin{document}

\maketitle

This document describes standard approaches for filtering and
estimation for quadrotors, created for the Udacity Flying Cars course.
We assume previous knowledge of probability and some knowledge of
linear algebra.  We do not assume previous knowledge of Kalman filters
or Bayes filters.  This document derives an EKF for various models of
drones in 1D, 2D, and 3D.  We use the EKF and notation as defined in
\citet{thrun2005probabilistic}.  We also give pseudocode for the Bayes
filter, the EKF, and the Unscented Kalman
filter~\citep{wan2000unscented}.  The motivation behind this document
is the lack of a step-by-step EKF tutorial that provides the
derivations for a quadrotor helicopter.  The goal of estimation is to
infer the drone's state (pose, velocity, accelleration, and biases)
from its sensor values and control inputs.  This problem is
challenging because sensors are noisy.  Additionally, because of
weight and cost issues, many drones have limited on-board computation
so we want to estimate these values as quickly as possible.  The
standard method for performing this method is the Extended Kalman
filter, a nonlinear extension of the Kalman filter which linearizes a
nonlinear transition and measurement model around the current state.
However the Unscented Kalman filter is better in almost every respect:
simpler to implement, more accurate to estimate, and comparable
runtimes.

\section{Averaging}

When performing estimation, the first thing one might think of is
averaging the sensor measurements.  We consider averaging for the 1D
case, where the range sensor value, $z_t$ is directly observing the
state (the height of the drone), $x_t$, both of which are scalars.
The notation $\hat{x_t}$ means our estimate of the true state $x_t$.
Then the average is:
\begin{align}
  \hat{x_t} &= \frac{1}{t} \Sigma_0^t z_t
\end{align}

\noindent However this form requires saving all sensor measurements from the beginning of time, requiring memory and computation that grows linearly in the number of observations.    Instead we rewrite the update recursively in terms of our previous estimate, $\hat{x}_{t-1}$, allowing us to transform $\hat{x}_{t-1}$ and $z_t$ to $\hat{x}$:
  \begin{align}
    \hat{x_t} &= \frac{1}{t} \left[ \hat{x}_{t-1} \times (t-1) + z_t \right].
  \end{align}

\noindent The recursive form, where the next estimate is written in
terms of the previous estimate, allows us to perform a constant time
update, and requires us to store a constant amount of information
about past sensor values (only the previous estimate, $\hat{x}_{t-1}$
- in other words, our estimate is Markov).  However this form updates
slowly when the drone's state changes.  Imagine the drone is
stationary for one minute, and then starts moving.  Then $\frac{1}{t}$
will be very small, and the estimate will take a long time to move to
the new value (longer and longer, the longer the drone is running).
Instead, we want an average that moves more quickly.  One way to
achieve a faster updating average is to simply average the old
estimate with the new sensor value:
\begin{align}
\hat{x_t} = \frac{\hat{x}_{t-1} + z_t}{2}
\end{align}

\noindent A more general formulation is a weighted average of the old
value and the new observation:
\begin{align}
\hat{x_t} = (\alpha)\hat{x}_{t-1} + (1-\alpha)z_t
\end{align}

This formulation allows the designer to tune the parameter $\alpha$,
which weights how much to weight the old estimate compared to the new
sensor value.

\section{Bayes Filter}

None of the above formulations update the estimate based on the
drone's movements.  Intuitively, if we have told the drone to go up,
for instance, then our belief about the drone's position should also
go up. The Bayes Filter gives us a way to incorporate motion
prediction into our state estimate.  First we {\em predict} the next
state, given a control input, the current state, and a model of how
the system evolves over time.  We do not maintain a point estimate but
rather a belief or distribution of the estimate.  Then, we revise our
prediction with an {\em update} from an observation.  The update
method takes the previous estimate from prediction, and an observed
sensor value.  It returns a new distribution that takes into account
the sensor value, using a model of how the sensor works.  Below you
will see pseudocode for the prediction and update steps for the
filter, following \citet{thrun2005probabilistic}.  In typical use one
would call predict after determining the control input, and update
after reading a sensor value.  The $\textsc{BayesFilter}$ function is
illustrative only; in real life, one might call predict in the sensor
callback for example.

\begin{algorithm}
  \caption{General Bayes Filter algorithm.  For specific filters such as the Kalman Filter or the particle filter, the representation for $bel$ and $\bar{bel}$ changes, and the corresponding mathematical updates take specific computational forms.}
  \begin{algorithmic}[1]
    \Function{Predict}{$bel(x_{t-1}), u_t, \Delta t$}
    \State $\bar{bel}(x_t) = \int p(x_t | u_t, x_{t-1}) bel(x_{t-1}) dx_{t-1}$
    \State \Return $\bar{bel}(x_t)$
    \EndFunction
    \Function{Update}{$\bar{bel}(x_{t}), z_t$}
    \State $bel(x_t) = \eta p(z_t | x_t) bel(x_t)$
    \State \Return $bel(x_t)$    
    \EndFunction
    \Function{BayesFilter}{}
    \State $u_t = \textsc{ComputeControl}(bel(x_{t-1}))$
    \State $\bar{bel}(x_t) = \textsc{Predict}(bel(x_{t-1}), u_t, \Delta t)$
    \State $z_t = \textsc{ReadSensor}()$
    \State $bel(x_t) = \textsc{Update}(\bar{bel}(x_t), z_t)$
    \EndFunction        
    \end{algorithmic}
\end{algorithm}

\section{E(KF)}

The Kalman Filter and Extended Kalman Filter make the assumption that
the distributions over belief state are Gaussian, represented as a
mean and covariance matrix. Compare to the Bayes' filter, the
distributions $bel$ and $\bar{bel}$ are represented as a mean and
covariance matrix.  The KF assumes that the transition and observation
models are linear, and can be defined by a matrix.  The EKF is the
extension to the nonlinear case, where we take use the Jacobian matrix
of the transition and observation functions to compute a point-wise
linear estimate, and then do the same updates as the Kalman Filter.
We define the Extended Kalman Filter (EKF) algorithm following
\citet{thrun2005probabilistic}.  We refactor it to include separate
$\textsc{Predict}$ and $\textsc{Update}$ methods and to use our
notation. We also unify the KF and EKF algorithm pseudocode.  The
transition or prediction covariance is $Q_t$; the measurement
covariance is $R_t$.  These matrices are often taken to be constant,
but also sometimes people change them over time depending on the
sensor model.

The EKF and KF are closely related.  For the KF, the matrix $G_t$ is
constant every iteration of the function and does not need to be
recomputed each time (except for $\Delta t$).  Another way to say it
is the implementation of $g'$ ignores its state input.  Similarly, for
the KF, the matrix $H_t$ is constant every iteration of the function
and does not need to be recomputed.  Another way to say it is the
function $h'$ ignores its state input.  For the EKF, these matrices
change each iteration, because it linearizes around the current
state.

\begin{algorithm}
  \caption{E(KF)  algorithm.  }  
  \begin{algorithmic}[1]
    \Function{Predict}{$\mu_{t-1}, \Sigma_{t-1}, u_t, \Delta t$}
    \State $\bar{\mu}_t = g(u_t, \mu_{t-1})$
    \State $G_t = g'(u_t, x_t, \Delta t)$
    \State $\bar{\Sigma}_t = G_t\Sigma_{t-1}G_t^T + Q_t$
    \State \Return $\bar{\mu}_t, \bar{\Sigma}_t$
    \EndFunction
    \Function{Update}{$\bar{\mu}_t, \bar{\Sigma}_t, z_t$}
    \State $H_t = h'(\bar{\mu}_t)$
    \State $K_t = \bar{\Sigma}_t H_t^T(H_t \bar{\Sigma}_t H_t^T + R_t)^{-1}$
    \State $\mu_t = \bar{\mu}_t + K_t(z_t - h(\bar{\mu}_t))$
    \State $\Sigma_t = (I - K_t H_t) \bar{\Sigma}_t$
    \State \Return $\mu_t, \Sigma_t$
    \EndFunction
    \Function{ExtendedKalmanFilter}{}
    \State $u_t = \textsc{ComputeControl}(\mu_{t-1}, \Sigma_{t-1})$
    \State $\bar{\mu}_t, \bar{\Sigma}_t = \textsc{Predict}(\mu_{t-1}, \Sigma_{t-1}, u_t, \Delta t)$
    \State $z_t = \textsc{ReadSensor}()$
    \State $\mu_{t}, \Sigma_t = \textsc{Update}(\bar{\mu}_t, \bar{\Sigma}_t, z_t)$
    \EndFunction    
  \end{algorithmic}
\end{algorithm}

\section{Unscented Kalman Filter}

The Unscented Kalman Filter~\citep{wan2000unscented} is similar to the
EKF in that it handles nonlinear transition models $g$ and measurement
models $h$.  However there are no Jacobians!  Instead of linearizing
around the current estimate, the Unscented Kalman Filter picks magic
``sigma points'' which are sample points chosen according to the
current state estimate and covariance.  Then these sigma points are
passed through the nonlinear transition or observation function, and
the sample mean and sample covariance of the sigma points is used to
construct a new Gaussian $\mu$ and $\sigma$.  The pseudocode here
follows \citet{kandepu2008applying} but removes the augmentation for
tracking moving prediction and measurement covariance, and uses our
notation.

Sigma points are computed using the matrix $S$ which is defined from
the covariance matrix, $\Sigma_t$.  $S_i$ denotes the $ith$ colum of
the matrix $S$.

\begin{align}
  S = \sqrt{\Sigma_t}
\end{align}

We define the sigma points $X_{i,t} \in X_t$ as follows:
\begin{align}
  X_{i,t} = \left\{  \begin{array}{lll}
    &= \mu_t,              & i=0\\
    &= \mu_t + \gamma S_i, & i=1,\dots,N\\
    &= \mu_t - \gamma S_{i-N}, & i=N+1,\dots,2N
  \end{array}
  \right.
  \label{eq:sigma}
\end{align}

Note that they are defined using the mean and covariance matrix of the
distributions; by picking several representative points $S_i$ away
from the mean, we can use a relatively small set of points to
represent the entire distribution.  The weights when computing the
sample mean, $w_i^m$ are:
\begin{align}
  w_i^m = \left\{  \begin{array}{lll}
  &= \frac{\lambda}{N + \lambda}            & i=0\\
  &= \frac{1}{2(N + \lambda)},              & i=1,\dots,2N\\
  \end{array}
  \right.
  \label{eq:weights}
\end{align}

The weights when computing the sample covariance, $w_i^c$ are:
\begin{align}
  w_i^c = \left\{  \begin{array}{lll}
  &= \frac{\lambda}{N + \lambda} + (1 - \alpha^2 + \beta)            & i=0\\
  &= \frac{1}{2(N + \lambda)},              & i=1,\dots,2N\\
  \end{array}
  \right.
  \label{eq:covweights}
\end{align}

The parameters are defined as: 
\begin{align}
  \gamma = \sqrt{N + \lambda}\\
  \lambda = \alpha^2(N + \kappa) - N
\end{align}

See \citet{kandepu2008applying} for tuning suggestions when
implementing the filter.

The pseudocode for the Unscented Kalman Filter is given in
Algorithm~\ref{alg:ukf}.  Compared to the Bayes' filter, we use the
computed sigma points to represent the distribution $\bar{bel}$.
However we use a mean and covariance to represent the distribution
$bel$ at the beginning and end of each iteration of the filter.

\begin{algorithm}
  \caption{Unscented Kalman Filter.  \label{alg:ukf}}  
  \begin{algorithmic}[1]

    \Function{ComputeSigmas}{$\mu_t, \Sigma_t$}
    \State \Return  $X_{0,t}, \dots, X_{2N,t}$ following Equation~\ref{eq:sigma}.
    \EndFunction
      
    \Function{Predict}{$\mu_{t-1}, \Sigma_{t-1}, u_t, \Delta t$}
    \State $X_{t-1} = \textsc{ComputeSigmas}(\mu_{t-1}, \Sigma_{t-1})$
    \State $\forall_{i=0}^{2N} \bar{X}_{i,t} = g(X_{i,t-1}, u_t, \Delta t)$
    \State \Return $\bar{X}_t$
    \EndFunction
    \Function{Update}{$\bar{X}_t, z_t$}
    \State $\bar{\mu}_t = \sum_{i=0}^{2N}(w_i^m \bar{X}_{i,t})$
    \State $\bar{\Sigma}_t = \sum_{i=0}^{2N} w_i^c (X_{i,t} - \bar{\mu}_t)(X_{i,t} - \bar{\mu}_t)^T+Q_t$
    \State $\forall_{i=1}^{2N} Z_{i,t} = h(\bar{X}_{i,t})$
    \State $\mu^z = \Sigma_{i=0}^{2N} w_i^m Z_{i,t}$
    \State $\Sigma_t^z = \Sigma_{i=0}^{2N} w_i^c (Z_{i,t} - \mu^z)(Z_{i,t} - \mu^z)^T+R_t$
    \State $\Sigma_t^{xz} = \Sigma_{0}^{2N} w_i^c (\bar{X}_{i,t} - \bar{\mu}_t)(Z_{i,t} - \mu^z)^T$
    \State $K_t = \Sigma_t^{xz} (\Sigma_t^z)^{-1}$
    \State $\mu_t = \bar{\mu_t} + K_t (z_t - \mu^z)$
    \State $\Sigma_t = \bar{\Sigma}_t - K_t \Sigma_t^z K_t^T$
    \State \Return $\mu_t, \Sigma_t$
    \EndFunction
    \Function{UnscentedKalmanFilter}{}
    \State $u_t = \textsc{ComputeControl}(\mu_{t-1}, \Sigma_{t-1})$
    \State $\bar{X}_t = \textsc{Predict}(\mu_{t-1}, \Sigma_{t-1}, u_t, \Delta t)$
    \State $z_t = \textsc{ReadSensor}()$
    \State $\mu_{t}, \Sigma_t = \textsc{Update}(\bar{X}_t, z_t)$
    \EndFunction    
  \end{algorithmic}
\end{algorithm}

\section{One Dimensional Quad}

To implement the filters on specific vehicles, we need to define the
state transition function, $g$ and the measurement model, $h$.  We
will do this three times for increasingly more realistic models of a
quadrotor.  First, we define a 1D quad model, where the quadroter is
moving in $z$, but not in $x$ or $y$.  It has one control input, the
downward pointing thrust, and a noisy range sensor.  The intention is
that this is identical to the 1D quad used in the controls lesson.
The state is then the position, $z$ and velocity, $\dot{z}$:

\begin{align}
  x_t = \left[\begin{array}{c}
      \dot{z}\\
      z
      \end{array}
      \right]
\end{align}

We define the control input as directly setting the accelleration, $\ddot{z}$:
\begin{align}
  u_t = \left[\begin{array}{c} \ddot{z} \end{array}\right]
\end{align}

\subsection{Transition Model}

Then we define a transition function $g(x_t, u_t, \Delta t)$ which returns a new $x_{t+1}$:

\begin{align}
  g(x_t, u_t, \Delta t) &= \left[\begin{array}{c}
      x_{t, \dot{z}} + u_{t,\ddot{z}}\times \Delta t\\ x_{t,z} + x_{t,\dot{z}} \times \Delta t \end{array} \right]\\
      &= \left[\begin{array}{cc}
          1 & 0\\
          \Delta t & 1 
    \end{array}\right]
      \left[\begin{array}{c} \dot{z}\\z\end{array}\right] +
      \left[\begin{array}{cc}
          \Delta t \\ 0
        \end{array}\right]
      \left[\begin{array}{c} \ddot{z} \end{array}\right]\\
      \intertext{We can rewrite it in terms of the $A_t$ and $B_t$ matrix, as in a conventional Kalman filter.  In this form we see that the control update is linear because it can be written in this form.}
      &=A_t x_t + B_t u_t
\end{align}

Then $x_t \in \mathbb{R}^2$ and $g(x_t, u_t, \Delta t) \in
\mathbb{R}^2$.  So $g'(x_t, u_t)$ is a $2\times2$ matrix, defined as
the partial derivative of $g$ with respect to $x_t$ for each component
in $x_t$.

\begin{align}
  g'(x_t, u_t, \Delta t) &= \left[\begin{array}{cccc}
      \frac{\partial}{\partial x_{t,\dot{z}}} g_{\dot{z}}(x_t, u_t, \Delta t) &
      \frac{\partial}{\partial x_{t,z}} g_{\dot{z}}(x_t, u_t, \Delta t)\\
      \frac{\partial}{\partial x_{t,\dot{z}}} g_z(x_t, u_t, \Delta t) &
      \frac{\partial}{\partial x_{t,z}} g_z(x_t, u_t, \Delta t)
    \end{array}\right]\\
  &=
  \left[\begin{array}{cccc}
      1&
      0\\
      \Delta t&
      1
    \end{array}\right]\\
\end{align}

Since this function is linear, the Jacobian is a constant matrix
except for $\Delta t$, and just reduces to the $A_t$ matrix.  

\subsection{Measurement Model}

Next we assume the drone has a range sensor pointed downwards at the
ground, at $z=0$.  Then $z_t$ is the range value, $r$:

\begin{align}
  z_t = \left[\begin{array}{c} r \end{array}\right]
\end{align}

Then we define a measurement function $h(x_t)$ which returns a new $z_t$:

\begin{align}
  h(x_t) &= \left[\begin{array}{c} x_{t,z} \end{array}   \right]\\
  &= \left[\begin{array}{cc} 0&1 \end{array}   \right] \left[\begin{array}{c}\dot{z}\\z\end{array}\right]\\
   &= C_t \left[\begin{array}{c}\dot{z}\\ z \end{array}\right]\\
   &= C_t x_t
\end{align}

Finally we define $h'(x_t)$, the Jacobian of $h$ with respect to
$x_t$.  The Jacobian is a $1\times2$ matrix.

\begin{align}
  h'(x_t) &= \left[\begin{array}{cccc}
      \frac{\partial}{\partial x_{t,\dot{z}}} h_r(x_t)&
      \frac{\partial}{\partial x_{t,z}} h_r(x_t)
    \end{array}   \right]\\
 &= \left[\begin{array}{cccc}
       0&
       1
    \end{array}   \right]  
\end{align}

Note that in this case the Jacobian does not depend at all on the
input $x_t$.  This is because this system is linear, so the EKF
linearization will boil back down to a regular Kalman filter.

\section{Two Dimensional Quad}
\label{sec:2d}
We define a 2D quad model where the quadrotor is operating at a fixed
height $z$.  It has a range sensor pointing sideways, and it can move
by rotating about its center, the angle $\phi$.  Additionally it can
move right and left in $y$.  Then we define the state $x_t$ as the
state transition function as follows:

\begin{align}
  x_t = \left[\begin{array}{c} \phi\\ \dot{y}\\ y\end{array}\right]
\end{align}

We define the control input as directly setting the angle, $\phi$. 
\begin{align}
  u_t = \left[\begin{array}{c} \phi \end{array}\right]
\end{align}

\subsection{Transition Model}

Then we define $g(x_t, u_t, \Delta t)$ and returns a new $x_{t+1}$:

\begin{align}
  g(x_t, u_t, \Delta t) = \left[\begin{array}{c} u_{t,\phi} \\ x_{t, \dot{y}} - \sin(x_{t,\phi}) \times \Delta t\\ x_{t,y} + x_{t,\dot{y}} \times \Delta t \end{array}\right]
\end{align}

Then $x_t \in \mathbb{R}^3$ and $g(x_t, u_t, \Delta t) \in
\mathbb{R}^3$.  So $g'(x_t, u_t)$ is a $3\times3$ matrix.

Finally, we take the partial derivative of $g$ with respect to $x_t$
for each component in $x_t$.

\begin{align}
  g'(x_t, u_t, \Delta t) &= \left[\begin{array}{cccc}
      \frac{\partial}{\partial x_{t,\phi}} g_\phi &
      \frac{\partial}{\partial x_{t,\dot{y}}} g_\phi &
      \frac{\partial}{\partial x_{t,y}} g_\phi\\
      \frac{\partial}{\partial x_{t,\phi}} g_{\dot{y}} &
      \frac{\partial}{\partial x_{t,\dot{y}}} g_{\dot{y}} &
      \frac{\partial}{\partial x_{t,y}} g_{\dot{y}}\\
      \frac{\partial}{\partial x_{t,\phi}} g_y &
      \frac{\partial}{\partial x_{t,\dot{y}}} g_y &
      \frac{\partial}{\partial x_{t,y}} g_y
    \end{array}\right]\\
&= \left[\begin{array}{cccc}
      \frac{\partial}{\partial x_{t,\phi}} u_{t,\phi} &
      \frac{\partial}{\partial x_{t,\dot{y}}} u_{t,\phi} &
      \frac{\partial}{\partial x_{t,y}} u_{t,\phi}\\
      \frac{\partial}{\partial x_{t,\phi}}  x_{t, \dot{y}} - \sin(x_{t,\phi}) \times \Delta t&
      \frac{\partial}{\partial x_{t,\dot{y}}} x_{t, \dot{y}} - \sin(x_{t,\phi}) \times \Delta t &
      \frac{\partial}{\partial x_{t,y}} x_{t, \dot{y}} - \sin(x_{t,\phi}) \times \Delta t\\
      \frac{\partial}{\partial x_{t,\phi}} x_{t,y} + x_{t,\dot{y}} \times \Delta t &
      \frac{\partial}{\partial x_{t,\dot{y}}}  x_{t,y} + x_{t,\dot{y}} \times \Delta t&
      \frac{\partial}{\partial x_{t,y}} x_{t,y} + x_{t,\dot{y}} \times \Delta t
    \end{array}\right]\\
&= \left[\begin{array}{cccc}
      0 &
      0 &
      0 \\
      -\cos(x_{t,\phi}) \Delta t &
      1 &
      0 \\
      0 &
      \Delta t &
      1
    \end{array}\right]
  \end{align}

\subsection{Measurement Model}

Next we assume the drone has a range sensor pointed sideways at a $\text{wall}_y=5$.  Then $z_t$ is the range value, $r$:
\begin{align}
  z_t = \left[\begin{array}{c} r \end{array}\right]
\end{align}

Then we define $h(x_t)$ and returns a new $z_t$:

\begin{align}
  h(x_t) = \left[\begin{array}{c} \frac{\text{wall}_y - x_{t,y}}{\cos(x_{t,\phi})} \end{array}   \right]
\end{align}

Finally we define $h'(x_t)$, the Jacobian of $h$ with respect to $x_t$.

\begin{align}
  h'(x_t) &= \left[\begin{array}{cccc}
      \frac{\partial}{\partial x_{t,\phi}} h_r(x_t)&
      \frac{\partial}{\partial x_{t,\dot{y}}} h_r(x_t)&
      \frac{\partial}{\partial x_{t,y}} h_r(x_t)
    \end{array}   \right]\\
  &= \left[\begin{array}{cccc}
      \frac{\left[\text{wall}_y - x_{t,y}\right]  \sin(x_{t,\phi})
        }
        {\cos^2(x_{t,\phi})}&
       0&
       \frac{-1}{\cos(x_{t,\phi})}
    \end{array}   \right]
\end{align}

Using the above math, we have implemented an EKF in Python for this
model, and showed it is able to estimate position and velocity using
the drone's simulated noisy range sensor.  

\section{Three Dimensional Quad}

\begin{table}
  \centering
  \begin{tabular}{cc}
    \toprule
    Variable & Description\\
    \midrule
    $u_t$ & The control input at time $t$.\\
    $x_t$ & The state at time $t$.\\
    $z_t$ & The observation at time $t$.\\
    $x_{t,\mbox{variable}}$ & \parbox{0.5\linewidth}{The scalar value of the state vector at the index corresponding to variable.  Similar notation for $u_t$ and $z_t$.}\\    
    $\Delta t$ & The elapsed time between updates in seconds.\\
    $Q_t$ & Transition model covariance\\
    $R_t$ & Measurement model covariance\\
    $\phi$ & Roll\\
    $\theta$ & Pitch\\
    $\psi$ & Yaw\\
    $R_{bg}$ & Rotation matrix from body to global\\
    $R_{gy}$ & Rotation matrix from global to yaw frame\\
    $x^y$& the $x$ coordinate in the yaw frame.\\
    $x$& The $x$ coordinate in the global frame.\\
    $x^b$& The $x$ coordinate in the body frame.\\
    $g(x_t, u_t, \Delta t)$ & The transition function. \\
    $h(x_t)$& The observation function.\\
    \bottomrule
  \end{tabular}
  \caption{Table of variables.}
\end{table}

The state will be position and velocity which we will estimate with the GPS.  We will use the magnetometer to estimate
yaw.  We will represent position and velocity in the global frame, and
yaw that goes from body to local in the state.  We would also then use
a multirate Kalman Filter~\citep{cristi2000multirate,
  quan2017introduction} to perform updating, with separate updates for
the GPS, the magnetometer, and the IMU (accelerometer plus rate gyro).

We track yaw as the heading from magnetic north.  So it is the reading
one would get if one reads from a compass.

\begin{align}
  x_t = \left[\begin{array}{c} 
      x\\
      y\\
      z\\
      \dot{x}\\
      \dot{y}\\
      \dot{z}\\
      \psi
      \end{array}\right]
\end{align}

Then $u_t$ is the acceleration in the body frame, where $\dot{\psi}$
is global frame yaw.

\begin{align}
  u_t &= \left[ \begin{array}{c}
      \ddot{x}^b \\
      \ddot{y}^b \\
      \ddot{z}^b \\
      \dot{\psi}
      \end{array} \right]
\end{align}

\subsection{Attitude Filter}
\label{sec:attitude}

\citet{markley2003attitude} gives an EKF in quaternions for attitude
estimation.  \citet{higgins1975comparison} compares complementary
filters to Kalman filters, in a way not specific to quads or attitude
estimation.  \citet{quan2017introduction} says to use either a
complementary filter or Kalman filter for attitude estimation and
gives very terse, hard to understand
math. \citet{johansen2017quadrotor} describes the MEKF model used in a
quadrotor with adaptive fading.  \citet{crassidis2007survey} gives a
survey of nonlinear attitude estimation including the MEKF.
\citet{nowicki2015simplicity} compares the complementary filter with
an EKF for attitude estimation on mobile phones.  They find that the
complementary filter is simpler to implement, but the EKF is able to
achieve in most cases better accuracy.  Both have comparable processor
loads.

We assume the state we are tracking is the vehicle's attitude, that is
roll $\phi$ and pitch, $\theta$.  Then the observation, $z_t$ consists
of the gyro angular velocity and pitch and roll angles as estimated
from the accelerometer.
\begin{align}
  z_t = \left[ \begin{array}{c}\theta\\\phi\\p\\q \end{array}\right]
\end{align}

The accelerometer estimates for $\theta$ and $\phi$ are in the global
frame, but the velocities from the gyro are in the body frame. Our
approach follows the Linear Complementary Filter from
\citet{quan2017introduction}.  We assume that $\theta$ and $\phi$ are
small, so that the turn rates measured by the gyro in the body frame
approximate the global turn rates.   In other words,
\begin{align}
  \left[\begin{array}{c}\dot{\phi}\\\dot{\theta}\\\dot{\psi}\end{array}\right]
  \approx
  \left[\begin{array}{c}p\\q\\r\end{array}\right]  
\end{align}

The state is the roll angle and
pitch angle:

\begin{align}
  x_t = \left[ \begin{array}{c}\theta\\\phi \end{array}\right]
\end{align}

\subsubsection{Linear Complementary Filter}

We define a linear complementary filter following
\citet{quan2017introduction}.  Here $\tau$ is a time constant and
$T_s$ is the filter sampling period:

\begin{align}
  \hat{\theta}_t = \frac{\tau}{\tau + T_s} \left( \hat{\theta}_{t-1} + T_s z_{t,\dot{\theta}} \right) + \frac{T_s}{\tau + T_s} z_{t,\theta}
\end{align}

Similarly for roll: 

\begin{align}
  \hat{\phi}_t = \frac{\tau}{\tau + T_s} \left( \hat{\phi}_{t-1} + T_s z_{t,\dot{\phi}} \right) + \frac{T_s}{\tau + T_s} z_{t,\phi}
\end{align}

We do not estimate the yaw with a complementary filter because we will
use the magnetometer and do it in the GPS.

The above math assumes that the angular velocity (which is in body
frame) can be used directly as angular rotation. 

\subsubsection{Nonlinear Complementary Filter}

For the nonlinear complementary filter, following
\citet{quan2017introduction} Section 9.1.3, we use the state to define
a quaternion, $q_t$, for the euler angles for $\phi$, $\theta$ and
$\psi$.  Then we can define $dq$ to be the quaternion that consists of
the measurement of the angular rates from the IMU in the body frame,
following Equation 84 in \citet{diebel2006representing}.  Using these
two, we can define a predicted quaternion, $\bar{q}_t$ as follows:

\begin{align}
  \bar{q}_t = dq * q_t 
\end{align}

Finally we can define $\bar{\theta}_t$ and $\bar{\phi}_t$ as follows:

\begin{align}
  \bar{\theta}_t = Pitch(\bar{q}_t)\\
  \bar{\phi}_t = Roll(\bar{q}_t)
\end{align}

Using these predicated estimates, we can compute the non-linear
complementary filter as above.
\begin{align}
  \hat{\theta}_t = \frac{\tau}{\tau + T_s} \left( \bar{\theta}_{t-1} + T_s z_{t,\dot{\theta}} \right) + \frac{T_s}{\tau + T_s} z_{t,\theta}
\end{align}

Similarly for roll: 

\begin{align}
  \hat{\phi}_t = \frac{\tau}{\tau + T_s} \left( \bar{\phi}_{t-1} + T_s z_{t,\dot{\phi}} \right) + \frac{T_s}{\tau + T_s} z_{t,\phi}
\end{align}

\subsection{Transition Model}

We define the transition function in terms of the rotation matrix
$R_{bg}$ which rotates from the body frame to the global frame.  As
described in \citet{diebel2006representing}, there are 12 different
orders one could perform the rotation; we follow the convention from
aerospace of using the $1,2,3$ order for roll, pitch, and yaw.

This matrix is defined as follows, taken from the transpose (or
inverse) of \citet{diebel2006representing}, equation 67.

\begin{align}
  R_{bg} = \left[\begin{array}{ccc}
      \cos\theta \cos \psi&
      \sin \phi \sin \theta \cos \psi - \cos \phi \sin \psi &
      \cos \phi \sin \theta \cos \psi + \sin \phi \sin \psi\\
      \cos \theta \sin \psi&
      \sin \phi \sin \theta \sin \psi + \cos \phi \cos \psi&
      \cos \phi \sin \theta \sin \psi - \sin \phi \cos \psi \\
      -\sin \theta&
      \cos \theta \sin \phi &
      \cos \theta \cos \phi
      \end{array}\right]
\end{align}

Then the transition function is:

\begin{align}
  g(x_t, u_t, \Delta t) &=
  \left[  \begin{array}{c}
      x_{t,x} +  x_{t,\dot{x}} \Delta t \\
      x_{t,y} +  x_{t,\dot{y}} \Delta t \\
      x_{t,z} + x_{t,\dot{z}} \Delta t\\
      x_{t,\dot{x}} \\
      x_{t,\dot{y}} \\
      x_{t,\dot{z}} - g \Delta t \\
      x_{t, \psi}\\
    \end{array}\right] + 
  \left[ \begin{array}{cccc}
      0&0&0&0\\
      0&0&0&0\\
      0&0&0&0\\
      R_{bg}[0:]&&&0\\
      R_{bg}[1:]&&&0\\
      R_{bg}[2:]&&&0\\
      0&0&0&1
      \end{array}
    \right]   u_t \Delta t
\end{align}

Then we take the Jacobian:
\begin{align}
  g'(x_t, u_t, \Delta t) &= \left [ \begin{array}{ccccccc}
      1 & 0 & 0 & \Delta t & 0 & 0 & 0\\
      0 & 1 & 0 & 0 & \Delta t & 0 & 0\\
      0 & 0 & 1 & 0 & 0 & \Delta t & 0\\
      0 & 0 & 0 & 1 & 0 & 0 & \frac{\partial}{\partial x_{t,\psi}} \left(x_{t,\dot{x}} + R_{bg}[0:]u_t[0:3] \Delta t\right)\\
      0 & 0 & 0 & 0 & 1  & 0 & \frac{\partial}{\partial x_{t,\psi}} \left(x_{t, \dot{y}} + R_{bg}[1:]u_t[0:3] \Delta t\right)\\
      0 & 0 & 0 & 0 & 0 & 1 & \frac{\partial}{\partial x_{t,\psi}} \left(x_{t, \dot{z}} + R_{bg}[2:]u_t[0:3] \Delta t\right)\\
      0 & 0 & 0 & 0 & 0 & 0 & 1\\ 
    \end{array}
    \right]\\
&= \left [ \begin{array}{ccccccc}
      1 & 0 & 0 & \Delta t & 0 & 0 & 0\\
      0 & 1 & 0 & 0 & \Delta t & 0 & 0\\
      0 & 0 & 1 & 0 & 0 & \Delta t & 0\\
      0 & 0 & 0 & 1 & 0 & 0 & R'_{bg}[0:]u_t[0:3] \Delta t\\
      0 & 0 & 0 & 0 & 1  & 0 & R'_{bg}[1:]u_t[0:3] \Delta t\\
      0 & 0 & 0 & 0 & 0 & 1 &  R'_{bg}[2:]u_t[0:3] \Delta t\\
      0 & 0 & 0 & 0 & 0 & 0 & 1
    \end{array}
    \right]
\end{align}

We define $R'_{bg}$ as $\frac{\partial}{\partial x_{t,\psi}}$, defined
  as Equation 71 from \citet{diebel2006representing}:
\begin{align}
R'_{bg} = \left[
  \begin{array}{ccc}
    -\cos \theta \sin \psi&
    -\sin\phi \sin \theta \sin \psi - \cos \phi \cos \psi&
    -cos \phi \sin \theta \sin \psi + \sin \phi \cos \psi\\
    \cos \theta \cos \psi&
    \sin \phi \sin \theta \cos \psi - \cos \phi \sin \psi&
    \cos \phi \sin \theta \cos \psi + \sin \phi \sin \psi\\
    0&0&0
  \end{array}
  \right]
\end{align}

\subsection{Measurement Model}

We provide measurement models for the GPS and Magnetometer. We use the IMU as a control input so do not provide a measurement model for it here.

\subsubsection{GPS}

We assume we get position and velocity from the GPS.  We considered
using heading from the GPS, but this does not take into account the
drone's orientation, only the direction of travel.  Hence we are
removing it from the observation.

\begin{align}
  z_t &= \left[ \begin{array}{c}
      x\\
      y\\
      z\\
      \dot{x}\\
      \dot{y}\\
      \dot{z}\\
      \end{array} \right]
\end{align}

Then the measurement model is: 
\begin{align}
  h(x_t) = \left[\begin{array}{c}
      x_{t,x}\\
      x_{t,y}\\
      x_{t,z}\\
      x_{t,\dot{x}}\\
      x_{t,\dot{y}}\\
      x_{t,\dot{z}}\\
    \end{array}\right]
\end{align}

Then the partial derivative is the identity matrix, augmented with a vector of zeros for $\frac{\partial}{\partial x_{t,\phi}} h(x_t)$: 
\begin{align}
  h'(x_t) = \left[\begin{array}{ccccccc}
      1&0&0&0&0&0&0\\
      0&1&0&0&0&0&0\\
      0&0&1&0&0&0&0\\
      0&0&0&1&0&0&0\\
      0&0&0&0&1&0&0\\
      0&0&0&0&0&1&0\\                  
      \end{array}\right]
\end{align}

\subsubsection{Magnetometer}

We assume we get a reading from the magnetometer reporting yaw in the
global frame.  (This measurement may need to be computed using roll
and pitch from the attitude filter and the mag vector.)
  
\begin{align}
  z_t=  \left[\begin{array}{c}
      \psi
      \end{array}\right]
  \end{align}

\begin{align}
  h(x_t) =  \left[\begin{array}{c}
      x_{t,\psi}
      \end{array}\right]
\end{align}

Again since this is linear, the derivative is a matrix of zeros and ones.

\begin{align}
  h'(x_t) = \left[\begin{array}{ccccccc}
      0&0&0&0&0&0&1
      \end{array}\right]
\end{align}

\subsection{Further Information}
Based on not tracking acceleration as part of the state, we will use
the accelerometer and gyro inputs as control inputs.  Note that
\citet{erdem2015fusing} describe using the accelerometer inputs in
the measurement or control input phases for an EKF over camera and
IMU, for all eight combinations.  They show that it is always better
to fuse both sensors in the measurement stage.  We would also then use
a multirate Kalman Filter~\citep{cristi2000multirate,
  quan2017introduction} to perform updating, with separate updates for
the GPS, magnetometer, and IMU.

  However this means that both acceleration and angular velocity appear
in the state vector, which we decided not to do in order to simplify
the math and implementation.  As one example, Ardupilot does it in
this way~\citep{ardupilotekf}.  The advantage is that it does not need
to connect as deeply to the control system.  Additionally, one could
even use different state transition models in the control compared to
the EKF.  One might make this decision for computational reasons, for
example.

\citet{erdem2015fusing} give the math for an EKF with an IMU and
camera (on a mobile device), showing the IMU and gyro measurements
treated as a prediction and measurement input (all eight
combinations).  They show that the best perforance is obtained using
the IMU as a measurement input.  However the difference in pose
accuracy estimation is around $\SI{1}{cm}$.  Note that the Ardupilot
open source code base~\citep{ardupilotekf} makes a different decision,
using the IMU as the control/prediction update, and that
\citet{bry2012state} describes this decision as a ``commonly-used
technique.''

We will use the North/East/Down frame where the positive $x$ gives the distance
along the surface of the earth in the direction of north; the $y$
coordinate gives the distance in the direction of $east$,
and $z$ is altitude, which is negative for distances above the surface
of the earth.

We define an intermediate frame, $p$, which is a quaternion
initialized from the roll, $\theta$, and the pitch, $\phi$, from the
complementary filter, and no yaw correction.  We define the global
frame, $q$, as a quaternion filled out with the roll, $\theta$, the
pitch, $\phi$, and the yaw, $\psi$, from the EKF.

\section{Conclusion}

We have provided equations for filtering for a quadrotor helicopter,
combining information and notation from
\citet{thrun2005probabilistic}, \citet{quan2017introduction} and other
sources.

\bibliographystyle{plainnat}
\bibliography{main}

\end{document}